\documentclass[10pt,twocolumn,letterpaper]{article}

\usepackage{cvpr}              
\usepackage{listings}
\usepackage{xcolor}
\usepackage{enumitem}
\usepackage{array}
\usepackage{amsmath}
\usepackage{amssymb}
\usepackage{booktabs}
\usepackage{dsfont}
\usepackage{graphicx}
\usepackage{makecell}
\usepackage{multirow}
\usepackage{pifont}
\usepackage[group-separator={,}]{siunitx}
\usepackage{tabularx}
\usepackage{xcolor}
\usepackage{caption}
\usepackage{soul}
\usepackage{changepage,threeparttable}
\usepackage{colortbl}
\usepackage{algpseudocode}
\usepackage{algorithm}
\usepackage{xspace}

\usepackage{mathtools}
\usepackage{xfrac}
\usepackage{url}

\usepackage{catchfile}
\usepackage{longtable}
\usepackage{tabu}
\usepackage{supertabular}
\usepackage{multicol}
\usepackage{enumitem}

\newcolumntype{Y}{>{\centering\arraybackslash}X}
\usepackage[accsupp]{axessibility}  

\definecolor{color_1}{RGB}{255,0,128}
\definecolor{color_2}{RGB}{0,128,128}
\definecolor{color_3}{RGB}{0,128,0}
\definecolor{color_4}{RGB}{128,0,0}
\definecolor{color_5}{RGB}{128,0,128}
\definecolor{cadetgrey}{RGB}{0.57, 0.64, 0.69}

\newif\ifarxiv
 \arxivtrue

\definecolor{cvprblue}{rgb}{0.21,0.49,0.74}
\definecolor{red}{rgb}{0.90,0.0,0.20}
\definecolor{green}{rgb}{0.49,0.76,0.00}
\definecolor{rulecolor}{HTML}{888888}
\definecolor{first}{HTML}{a9c98f}
\definecolor{second}{HTML}{cadeba}
\definecolor{third}{HTML}{eaf2e3}
\definecolor{beige}{HTML}{f2f0e9}
\definecolor{sunny}{HTML}{fffce8}
\newcommand{\rgbx}{RGB$\xleftrightarrow{}$X\xspace}
\newcommand{\prism}{PRISM\xspace}
\newcommand{\method}{ReasonX\xspace}
\newcommand{\prismx}{PRISM-\textbf{X}\xspace}
\newcommand{\marigoldx}{Marigold-\textbf{X}\xspace}
 
\newcommand{\cI}{\cellcolor{first}}
\newcommand{\cII}{\cellcolor{second}}
\newcommand{\cIII}{\cellcolor{third}}
\newcommand{\cg}{\cellcolor{lightgray}}
\newcommand{\ours}{\cellcolor{beige}}

\newcommand{\impr}{\cellcolor{sunny}}
\usepackage[pagebackref,breaklinks,colorlinks,allcolors=cvprblue]{hyperref}
\usepackage{amsmath}
\usepackage{colortbl}
\usepackage{makecell}

\arrayrulecolor{rulecolor}
\renewcommand{\arraystretch}{1.1}
\usepackage[T1]{fontenc} 
\usepackage{iftex}
\ifLuaTeX
\protected\def\pdfmapline {\pdfextension mapline }
\fi
\pdfmapline{+goodlife < Goodlife.ttf <T1-WGL4.enc}

\def\paperID{XXXX} 
\def\confName{CVPR}
\def\confYear{2026}

\title{\method: MLLM-Guided Intrinsic Image Decomposition}

\author{Alara Dirik$^{1}$, Tuanfeng Wang$^{2}$, Duygu Ceylan$^{2}$, Stefanos Zafeiriou$^{1}$, Anna Frühstück$^{2}$
\\
$^1$Imperial College London \hspace{5em} 
$^2$Adobe Research
}

\begin{document}
\twocolumn[{%
\renewcommand\twocolumn[1][]{#1}%
\maketitle
\begin{center}
    \centering
    \captionsetup{type=figure}
    \vspace{-1.5em}
    \includegraphics[width=\textwidth]{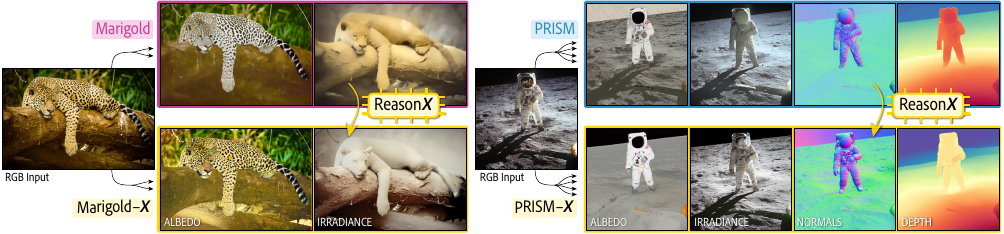}
    \vspace{-1.8em}
    \captionof{figure}{We propose Reason\textbf{X}, a novel framework for MLLM-guided improvement of intrinsic decomposition models via relative intrinsic judgments on RGB input images.} 
\end{center}%
}]


\begin{abstract}
Intrinsic image decomposition aims to separate images into physical components such as albedo, depth, normals, and illumination. While recent diffusion- and transformer-based models benefit from paired supervision from synthetic datasets, their generalization to diverse, real-world scenarios remains challenging. We propose \method, a novel framework that leverages a multimodal large language model (MLLM) as a perceptual \emph{judge} providing \emph{relative} intrinsic comparisons, and uses these comparisons as GRPO rewards for fine-tuning intrinsic decomposition models on unlabeled, in-the-wild images. Unlike RL methods for generative models, our framework aligns \emph{conditional} intrinsic predictors by rewarding agreement between the judge's relational assessments and analytically derived relations from the model's outputs. \method is model-agnostic and can be applied to different intrinsic predictors. Across multiple base architectures and modalities, \method yields significant improvements, including 9-25\% WHDR reduction on IIW albedo and up to 46\% depth accuracy gains on ETH3D, highlighting the promise of MLLM-guided comparative supervision to bridge low- and high-level vision reasoning.
\end{abstract}
\vspace{-2em}
\section{Introduction}
Recovering intrinsic scene properties such as albedo, surface normals, depth, and illumination from a single RGB image is a fundamental inverse problem in computer vision. Remarkable progress has been made on this task with the adoption of diffusion models and vision transformers as backbone architectures~\cite{Careaga2024ColorfulDI, Dirik2025PRISMAU, Kocsis2024IntrinsicID, Ke2025MarigoldAA, Fu2024GeoWizardUT}. Recent methods have demonstrated impressive capabilities in intrinsic scene understanding, achieving high-quality decompositions on complex indoor scenes, and enabling a variety of downstream applications such as relighting and material editing. These advances have been driven not only by sophisticated model architectures, but also by the development of large-scale training frameworks~\cite{Zeng2024RGBXID} that leverage multiple heterogeneous datasets~\cite{InteriorVerse,roberts2021hypersim}. 

\begin{figure*}[t]
\vspace{-1em}
    \centering
    \includegraphics[width=\textwidth]{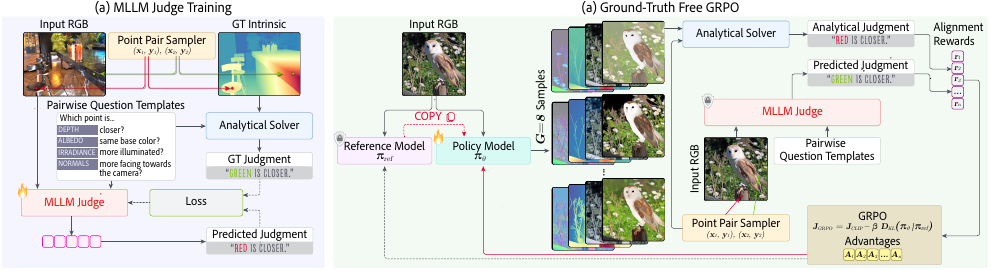}
    \caption{
        \textbf{Overview of our \method framework.} (a) We fine-tune an MLLM to judge relative intrinsic properties from RGB images using sampled point pairs. 
        (b) The frozen judge then provides rewards within a GRPO loop to refine an intrinsic decomposition model $\pi$: for each RGB image, we generate a group of $G=8$ samples, query the judge across point pairs and modalities, and compute group-relative rewards to update $\pi$ without ground-truth intrinsics.
    }
    \vspace{-1.5em}
    \label{fig:pipeline}
\end{figure*}

However, a fundamental bottleneck remains: the reliance on paired synthetic datasets with ground-truth intrinsic labels. State-of-the-art methods depend heavily on synthetic data rendered from physically-based simulation engines~\cite{InteriorVerse, roberts2021hypersim, li2021openrooms, Li2023MatrixCityAL}, which, while photorealistic, are inherently limited in scope and diversity. These datasets typically focus on specific domains (e.g., indoor scenes) and fail to capture the full complexity of in-the-wild imagery. Creating comprehensive ground truth for intrinsic properties across diverse real-world scenarios remains prohibitively expensive and technically challenging. Consequently, models trained primarily on synthetic data often struggle to generalize to images outside their training distribution, limiting their practical applicability.

Meanwhile, multimodal large language models (MLLMs) have emerged as powerful foundational models with extensive visual and semantic understanding capabilities~\cite{Bai2023QwenVLAF, Zhu2025InternVL3EA}. Recent work has demonstrated that MLLMs achieve impressive performance on diverse visual reasoning tasks, even approaching or surpassing specialized models in certain domains~\cite{li2024enhancing,chen2024lion}. This raises a compelling question: can we leverage the priors from pretrained MLLMs to overcome the generalization bottleneck in intrinsic image decomposition?
We take inspiration from recent findings that MLLMs excel at \emph{relative} spatial reasoning~\cite{Zhan2025Open3DVQAAB, Wu2025SpatialMLLMBM, Xu2025MultiSpatialMLLMMS}. This observation aligns with principles of human perception as people are naturally better at making comparative judgments (e.g., \textit{'which point is closer?'}, \textit{'do these regions share the same material?'}) than absolute measurements~\cite{Bell2014Intrinsic}. Building on this insight, we propose to fine-tune pre-trained MLLMs as \emph{judges} that evaluate the quality of intrinsic decompositions through pairwise comparisons. 

Our key insight is to treat these relative intrinsic judgments as a supervision signal. Rather than learning intrinsics from absolute labels or heuristic losses, we train an MLLM to reason about relative intrinsic comparisons by sampling point pairs on images and asking questions about their intrinsic properties (e.g., relative depth, material similarity, lighting differences). We then employ this judge as a reward function within a Group Relative Policy Optimization (GRPO) framework~\cite{Shao2024DeepSeekMathPT}. Because intrinsic decomposition is tightly conditioned on the input RGB image and therefore largely deterministic, we introduce exploration by rewarding consistency between the judge’s comparative assessments and analytical relations computed from the model’s predicted intrinsics. 
During fine-tuning of the base intrinsic decomposition model, we sample multiple point pairs per image and compute alignment rewards by comparing the judge's assessment with predicted intrinsics. This approach requires no ground-truth intrinsic images during the base model fine-tuning, we only need RGB images and an MLLM judge trained to reason about intrinsic properties through relative comparisons. 

We evaluate the effectiveness of our framework by coupling our MLLM-based judge with different base intrinsic methods, including flow matching models (PRISM~\cite{Dirik2025PRISMAU}) and diffusion-based models (Marigold IID Lighting v1.1~\cite{Ke2025MarigoldAA}). The supervision from the MLLM judge enables training of the base model on diverse in-the-wild images, achieving 9–25\% WHDR reduction on the IIW albedo benchmark and up to 46\% depth accuracy improvements on ETH3D.  We also establish an out-of-distribution dataset and show that our method significantly improves the base model's generalization to challenging real-world scenarios. 
Our main contributions include:
\begin{itemize}
\item An MLLM-based judge that reasons about intrinsic image decompositions through pairwise comparisons, 
\item A GRPO-based approach that uses the MLLM judge as a reward signal to fine-tune base intrinsic models on diverse and unlabeled images,
\item Improved generalization to in-the-wild scenes across different intrinsic modalities (albedo, depth, normals, and irradiance) and multiple base models.
\end{itemize}
\vspace{-0.5em}
\section{Related Work}
\begin{figure*}[t]
    \centering
    \includegraphics[width=1\linewidth]{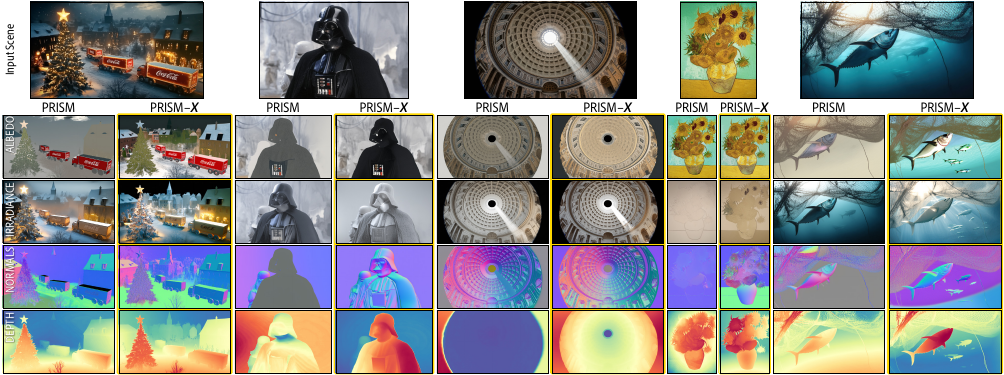}
    \vspace{-2em}
    \caption{Intrinsic decomposition samples on challenging out-of-distribution images. Our PRISM-\textbf{X} significantly improves its base model PRISM across all intrinsic channels with respect to decomposition quality and in-the-wild generalization performance.} 
    \label{fig:ood}
    \vspace{-1em}
\end{figure*}

\paragraph{Intrinsic Image Decomposition Models.}
Early intrinsic decomposition methods differ mainly in how they model the interaction between reflectance, geometry, and illumination, ranging from explicit physical priors~\cite{Land1971LightnessAR} to data-driven reflectance–illumination separation~\cite{narihira2015learning, Baslamisli2018Joint}. Among recent advances, Ordinal Shading~\cite{OrdinalShading} presents a dual-stage convolutional framework that enforces scale- and shift-invariant ordering within shading predictions. By focusing on relative rather than absolute intensity, it preserves global coherence and fine detail, demonstrating the benefits of ordinal constraints for intrinsic decomposition. Our approach extends this principle beyond shading, using a multimodal large language model (MLLMs) as a perceptual judge to evaluate relative relationships across arbitrary intrinsic modalities-such as depth, reflectance, or illumination. Through this generalized ordinal reasoning and GRPO-based optimization, our method enables cross-modal fine-tuning of existing models without paired supervision.

Complementary to these advances, vision transformers and diffusion-based architectures have enabled multi-modal intrinsic prediction with improved consistency~\cite{Dirik2025PRISMAU, Ke2025MarigoldAA, Fu2024GeoWizardUT, Careaga2024ColorfulDI}. Frameworks such as PRISM~\cite{Dirik2025PRISMAU} and Marigold~\cite{Ke2025MarigoldAA} jointly predict intrinsic properties, demonstrating strong performance on synthetic datasets, while our method provides an orthogonal route toward generalization through MLLM-guided, order-aware refinement. For a broader overview of intrinsic decomposition methods, we refer readers to recent surveys~\cite{Garces2021ASO}.
\vspace{-0.8em}

\vspace{-.5em}
\paragraph{Synthetic Data and Generalization.}
Due to the scarcity of real-world intrinsic labels, many methods rely on physically-based rendering datasets such as HyperSim~\cite{roberts2021hypersim}, InteriorVerse~\cite{InteriorVerse}, OpenRooms~\cite{li2021openrooms}, and MatrixCity~\cite{Li2023MatrixCityAL}. While photorealistic, these datasets are typically biased toward indoor environments, limiting generalization to natural scenes. To address this, prior work has explored cross-domain learning, semi-supervised approaches, and auxiliary modalities such as RGBX supervision~\cite{Zeng2024RGBXID, Li2023MatrixCityAL}.

\vspace{-1.2em}
\paragraph{Multimodal Large Language Models for Visual Reasoning.}
Multimodal large language models (MLLMs)~\cite{Bai2023QwenVLAF, Zhu2025InternVL3EA} combine vision encoders with large-scale language backbones, enabling reasoning about spatial and semantic relationships in images. Initially applied to high-level tasks such as captioning and visual question answering, MLLMs have recently been adapted for geometric reasoning in 3D scenes~\cite{Zhan2025Open3DVQAAB, Wu2025SpatialMLLMBM, Xu2025MultiSpatialMLLMMS, GeoPQA2025, VisualJigsaw2025}. These models excel at relative and comparative judgments, although absolute metric estimation remains challenging.

\vspace{-1.2em}
\paragraph{MLLM-Based Feedback and Reward-Guided Learning.}
Recent works investigate using MLLMs as verifiers or reward models for feedback-guided generation~\cite{Zhang2025ThinkBY, Luo2025DualProcessIG, Bai2024RewardBench, RLVLMF2024, CodeAsReward2024, VisionLanguageReward2023}, extending reinforcement learning from human feedback~\cite{Christiano2017Deep} to multimodal tasks. Techniques like Group Relative Policy Optimization (GRPO)~\cite{Shao2024DeepSeekMathPT} have been used to align generative models via learned critics~\cite{Liu2025FlowGRPOTF}.

\vspace{-1.2em}
\paragraph{Relative Reasoning and Comparative Learning.}
Relative supervision has been recognized as a robust alternative to absolute labels in vision tasks. Works on ordinal depth estimation~\cite{Chen2016Single, zoran2015learning} and pairwise reflectance comparisons~\cite{Bell2014Intrinsic} illustrate how comparative cues provide stronger structural constraints and improved generalization. Recent MLLM research demonstrates strong spatial and relational judgment capabilities~\cite{Zhan2025Open3DVQAAB, Wu2025SpatialMLLMBM, GeoPQA2025}, suggesting a natural connection between comparative supervision and multimodal reasoning.

\vspace{-.5em}
\section{Method} \label{sec:method}
Our goal is to fine-tune a base intrinsic image decomposition model with real, unlabeled images. To achieve this, we train a multimodal large language model (MLLM) to act as a perceptual judge that makes \emph{relative intrinsic comparisons}, and then fine-tune the base model to satisfy these comparisons through Group Relative Policy Optimization (GRPO). As illustrated in \Cref{fig:pipeline}, our \method framework consists of two components: (i) a pretrained intrinsic model that predicts multiple physical modalities from an RGB image, and (ii) an MLLM judge that evaluates these predictions through point-pair comparisons.

\vspace{-1em}
\paragraph{Base Models.}
\method is flexible and can be applied to various intrinsic decomposition models. 
We demonstrate this using two complementary architectures: 
PRISM~\cite{Dirik2025PRISMAU}, a diffusion transformer trained under the rectified flow formulation for text- and image-conditioned RGBX generation, and Marigold~\cite{Ke2025MarigoldAA} IID Lighting v1.1, a diffusion-based model for joint albedo and irradiance estimation. 

\subsection{MLLM Judge: Relative Intrinsic Reasoning}

\begin{figure}[t]
    \centering
    \includegraphics[width=1\linewidth]{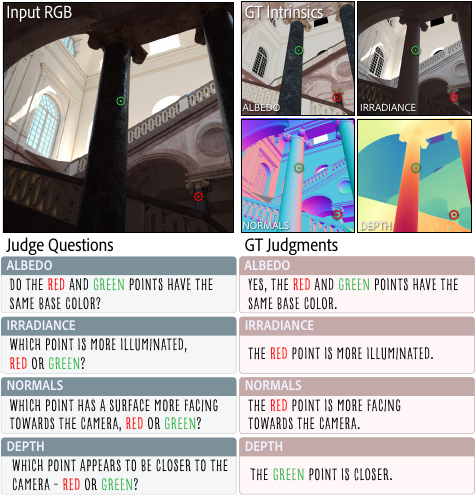}
    \vspace{-1.5em}
\caption{Our MLLM judge is trained on synthetic data to make relative intrinsic judgments from point-pair annotated RGB images and modality-specific questions. Ground-truth answers are derived from the corresponding intrinsic maps.}
    \label{fig:judge}
    \vspace{-2.0em}
\end{figure}

We adopt a multimodal language model (MLLM) with image understanding capability to perform pairwise intrinsic judgments on RGB images. More specifically, we fine-tune InternVL2.5-4B~\cite{Chen2024ExpandingPB} on synthetic RGB images from the base model's training set, where ground-truth intrinsics are available. For each training image, we sample point pairs $(x_1, y_1), (x_2, y_2)$, overlay colored visual markers, and formulate modality-specific relative questions (see \Cref{fig:judge}) such as \textit{"which point is closer?"} or \textit{"do these points share the same material?"}. This focuses the learning task on \emph{comparative} reasoning, which we find transfers far more reliably to real images than absolute predictions (see supplementary for analysis). Ground-truth comparisons are derived analytically from corresponding synthetic intrinsic maps as follows:

\begin{itemize}
    \item \textbf{Depth:} direct comparison of scalar depth values; pairs with minimal difference are excluded to avoid ambiguity.
    \item \textbf{Normals:} assuming the camera is looking at the $+z$ axis; comparison of front-facingness via the normal’s $z$-component. 
    \item \textbf{Irradiance:} comparison of luminance-albedo ratios under Lambertian assumptions; the point with the higher ratio is labeled as more illuminated. 
    \item \textbf{Albedo:} thresholded perceptual color differences. 
\end{itemize}

Once trained, the MLLM judge remains fixed. For a predicted intrinsic map $\mathbf{I}_m$, we sample $N$ point pairs on the RGB input, overlay visual markers, and query the frozen judge with a modality-specific question to get a relative judgment. We then derive analytic predictions from $\mathbf{I}_m$, and compute their agreement as our alignment reward:

\begin{equation}
r(\mathbf{I}_{\text{RGB}}, \mathbf{I}_m)
= \frac{1}{N} \sum_{i=1}^N 
\mathbb{1}\!\left[
\mathrm{MLLM}(\mathbf{I}_{\text{RGB}}^{(i)}, q_m)
=
g_m(\mathbf{I}_m, \mathbf{p}_i)
\right],
\end{equation}

where $\mathbf{I}_{\text{RGB}}^{(i)}$ is the RGB image with markers at point pair $\mathbf{p}_i$, $q_m$ is the modality-specific query, and $g_m$ denotes the deterministic relation computed from the predicted intrinsic map. This produces a reward signal grounded in both the model’s physical structure and the judge’s comparative perception. We next describe how this reward is used to fine-tune the base intrinsic model.

\subsection{Intrinsic-GRPO: Learning from Relative Rewards}
Fine-tuning an intrinsic decomposition model on real images is challenging as the task is \emph{RGB-conditioned and largely deterministic}, leaving little room for the exploration required by policy-gradient methods. We therefore formulate the fine-tuning of the base intrinsic decomposition model as a reinforcement learning problem. Specifically, we perform such a fine-tuning on a set of real images where no ground truth annotations exist. We obtain intrinsic predictions from the current base model, compute rewards for these predictions using the MLLM-based judge, and back-propagate this reward as a supervision signal.

\vspace{-.5em}
\paragraph{Injecting Exploration.}
Both PRISM and Marigold generate intrinsic predictions through a deterministic, image-conditioned denoising trajectory. 
In both cases, the model applies 
a learned update $f_\theta(\mathbf{x}_t, t, \mathbf{c})$ to the latent state $\mathbf{x}_t$ at each timestep $t$ given the condition $\mathbf{c}$. To enable exploration, we follow Flow-GRPO~\cite{Liu2025FlowGRPOTF} and introduce a small stochastic term during sampling, converting the deterministic trajectory into a lightly stochastic one via the Euler–Maruyama update:
\begin{equation}
\mathbf{x}_{t+\Delta t}
= \mathbf{x}_t
+ f_\theta(\mathbf{x}_t, t, \mathbf{c})\,\Delta t
+ \sigma_t \sqrt{\Delta t}\,\boldsymbol{\epsilon},
\end{equation}
where $\boldsymbol{\epsilon}\!\sim\!\mathcal{N}(\mathbf{0},\mathbf{I})$ and $\sigma_t$ controls the noise level. We note that independent noise vectors yield multiple plausible intrinsic predictions for the same RGB image, which is crucial for comparative, group-based optimization in an otherwise deterministic setting.

\vspace{-.5em}
\paragraph{Group-based reward optimization.}
During policy learning on real images, we keep the judge MLLM and the VAE of the base model frozen. For each real RGB image, we randomly choose a target modality $m$, generate a group of $G$ intrinsic predictions using independent noise, compute a reward for each sample using the MLLM judge, and update the model using GRPO. Given group rewards $\{r_i\}_{i=1}^G$, we compute normalized advantages as
\[
\hat{A}_i = \frac{r_i - \mu_G}{\sigma_G},
\]
where $\mu_G$ and $\sigma_G$ are the group mean and standard deviation. GRPO optimizes a clipped PPO-style objective~\cite{Shao2024DeepSeekMathPT} with a KL regularizer that constrains the updated policy to remain close to a frozen reference copy of the pretrained model:
\begin{equation}
\begin{aligned}
\mathcal{J}(\theta)
= \mathbb{E}_{\pi_\theta}\Big[
&\min\!\left(\rho_t \hat{A},
\,\mathrm{clip}(\rho_t,1-\epsilon,1+\epsilon)\hat{A}\right) \\
&\quad -\, \beta\, D_{\mathrm{KL}}(\pi_\theta \,\|\, \pi_{\mathrm{ref}})
\Big],
\end{aligned}
\end{equation}
where $\rho_t=\pi_\theta(\mathbf{x}_{t-1}|\mathbf{x}_t)/\pi_{\theta_{\text{old}}}(\mathbf{x}_{t-1}|\mathbf{x}_t)$ is the importance ratio. We note that the KL regularization helps prevent reward hacking~\cite{Liu2025FlowGRPOTF}, such as collapsing to near-constant intrinsics, and ensures that improvements reflect genuine alignment with comparative judgments. Under the stochastic update above, the transition distribution is Gaussian, allowing the KL divergence to be computed in closed form by comparing the model and reference velocity fields.

Our formulation adapts GRPO to the intrinsic decomposition setting by combining RGB-conditioned exploration with structured, relative rewards from the MLLM judge. Using our \method framework, the base intrinsic decomposition model is refined to satisfy physics-guided relational constraints directly on real images, without requiring any ground-truth intrinsic maps. Unlike prior uses of GRPO in image generative settings~\cite{Liu2025FlowGRPOTF}, our method targets deterministic, RGB-conditioned intrinsic prediction, where the reward is derived from modality-specific relational comparisons rather than preference scores.

\subsection{Implementation Details}
We fine-tune each base model on a dataset of 10,000 real RGB images from the COCO training set~\cite{Lin2014MicrosoftCC}. At each iteration, we sample $N=40$ point pairs and use $T=15$ denoising steps (inference uses $T=50$). PRISM is conditioned with an empty text prompt. We optimize both base models using AdamW with a learning rate of $10^{-5}$, cosine annealing schedule, and gradient clipping at norm 1.0. The group size is set to $G=8$, the SDE noise level to $a=0.7$. We train our models on 6 H100 GPUs for 3 epochs.

\section{Evaluation}
\begin{figure}[ht]
    \centering
    \includegraphics[width=0.9999\linewidth]{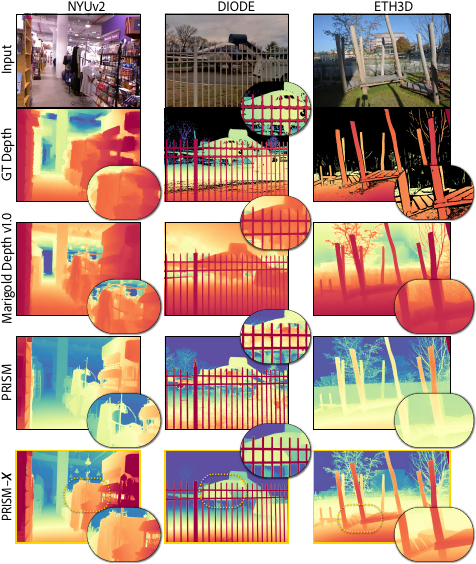}
    \vspace{-2em}
    \caption{Depth estimation comparisons of our \prismx with its base model \prism and SOTA baseline method Marigold Depth v1.0 on samples from the NYUv2, DIODE and ETH3D datasets. \prismx performs significantly better on challenging images.} 
    \label{fig:depth_comparison}
    \vspace{-1em}
\end{figure}

\begin{figure}[t]
    \centering
    \includegraphics[width=1\linewidth]{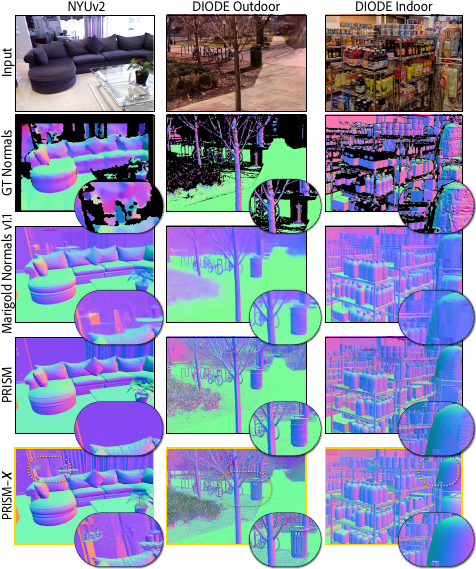} 
    \vspace{-2em}
    \caption{Normals estimation comparisons of our \prismx with its base model \prism and SOTA baseline method Marigold Normals v1.1 on NYUv2 and DIODE indoor and outdoor samples.} 
    \label{fig:normals_comparison}
    \vspace{-1.5em}
\end{figure}

\method is a generic framework that can be used in conjunction with different base intrinsic decomposition models. Hence, we center our evaluations to showcase the relative performance gains achieved through our \method, both qualitatively and quantitatively, across multiple modalities and datasets. In particular, we apply \method on top of two representative base models - PRISM and Marigold IID Lighting v1.1 \cite{Ke2025MarigoldAA} (a Marigold variant that jointly estimates albedo, irradiance and residual images) - resulting in enhanced variants denoted as \prismx and \marigoldx, respectively. To further contextualize these improvements, we include an additional MLLM baseline where we fine-tune OmniGen2 \cite{Wu2025OmniGen2ET}, a versatile MLLM with image generation and editing capabilities to generate intrinsic modalities (e.g. depth, albedo) given an RGB image and a text prompt cue denoting the task (e.g. “convert this image to a depth map.”). We use the same synthetic dataset that we train our MLLM judge for this fine-tuning. While this baseline provides competitive performance, it often is inferior to \method variants. In addition, we report results from task-specific state-of-the-art methods to position our models within the broader literature, noting that many of these SOTA approaches are trained on substantially larger or domain-specific datasets. While this may limit direct comparability, we observe that our method is often able to match the performance of specialized models. We refer to the supplementary material for a more comprehensive evaluation including additional baseline methods.

\textbf{Evaluation Metrics and Datasets.}
We use various modality specific datasets and metrics for a thorough comparison. For albedo decomposition, we perform our comparisons on the real IIW~\cite{Bell2014Intrinsic} and MAW~\cite{Wu2023MeasuredAI} datasets. 
We follow the experiment protocol in \citet{OrdinalShading, Careaga2024ColorfulDI} and report the mean weighted human disagreement rate (WHDR), intensity and chromaticity metrics. 

For depth estimation, we perform zero-shot evaluation on the NYU-v2 \cite{Silberman2012IndoorSA} (indoor) and ETH3D~\cite{Schps2017AMS} (mixed indoor and outdoor) datasets and report Absolute Relative Error (AbsRel) and Threshold Accuracy ($\delta_1$) metrics following previous work. For surface normals estimation, we report the mean angular error and the percentage of pixels with angular error lower than 11.5° on the NYU-v2 and DIODE~\cite{diode_dataset} datasets. Unless specified otherwise, we perform all experiments on the designated test set of each dataset. 

\subsection{Intrinsic Decomposition Performance}
\textbf{Albedo Estimation.} 
We start by evaluating our method on two real datasets with respect to the albedo estimation quality and report the numbers in \Cref{tab:iiw,tab:maw}, respectively. IIW~\cite{Bell2014Intrinsic} provides pairwise human annotations of albedo brightness between sparsely sampled pixels. The MAW dataset~\cite{Wu2023MeasuredAI} consists of $\sim$850 indoor images and measures albedo within specific masked regions of the image. For MAW, we report the intensity and chromaticity of the albedo estimates. 

\begin{table}[h]
\caption{WHDR error of albedo estimates on the IIW dataset~\cite{Bell2014Intrinsic} for our base models, their \method variants and select SOTA baselines. CRefNet~\cite{luo2023crefnet}, a non-competing method trained on IIW, is denoted with gray. Our \method variants significantly outperform their respective base models (9-25\% improvement in WHDR 10\%) and achieve SOTA results compared to highest performing zero-shot methods.}
\vspace{-1.5em}
\label{tab:iiw}
\begin{center}
\scalebox{0.8}{
\begin{tabularx}{1.155\linewidth}{lcc}
\toprule
\textbf{Method} & \textbf{WHDR 10\%}~$\downarrow$ & \textbf{WHDR 20\%}~$\downarrow$ \\
\midrule
CRefNet \cite{luo2023crefnet}          & \cg 12.8    & \cg 10.8    \\
\midrule
\citet{InteriorVerse}  & 34.7             & 24.1             \\
\citet{OrdinalShading}  & 24.8             & 19.2             \\
\citet{Kocsis2024IntrinsicID}   & 26.1             & 20.7             \\
\rgbx \cite{Zeng2024RGBXID}     & 23.6      & 21.1             \\
\citet{Careaga2024ColorfulDI} & 16.8    & 15.6             \\
OmniGen2 \cite{Wu2025OmniGen2ET} & 17.8    & 14.3             \\
\midrule
Marigold (base variant) \cite{Ke2025MarigoldAA} & \cIII 16.7 & \cIII 14.8 \\
\ours \marigoldx & \cII \textbf{15.2} & \cII 14.0 \\
\multicolumn{1}{l}{\impr\textit{Improvement}} &\impr\textit{+9.0\%} & \impr\textit{+5.4\%} \\
\midrule
PRISM \cite{Dirik2025PRISMAU}       & 17.2       & 15.9      \\
\ours \prismx             & \cI \textbf{12.9}            & \cI \textbf{11.9}       \\
\multicolumn{1}{l}{\impr\textit{Improvement}} &\impr\textit{+25.0\%} & \impr\textit{+25.2\%} \\
\bottomrule
\end{tabularx}
}
\end{center}
\vspace{-.5em}
\end{table}

\begin{figure*}[t]
    \centering
    \includegraphics[width=1\linewidth]{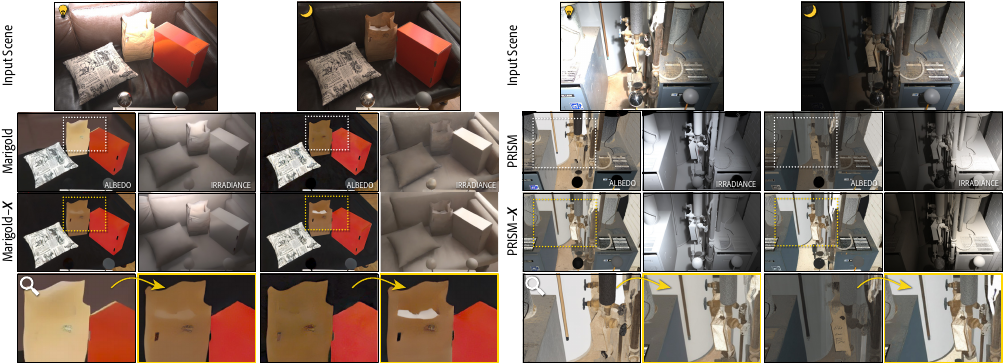}
    \vspace{-1.8em}
    \caption{Effect of \method on albedo and irradiance decomposition under varying lighting conditions. Note how both models are able to significantly improve their stability under very difficult lighting settings, such as these over/underexposed images.} 
    \label{fig:consistent-decomposition}
    \vspace{-1.2em}
\end{figure*}

\begin{table}
\vspace{-0.5em}
\caption{Intensity and chromaticity of albedo estimates on the MAW dataset~\cite{Wu2023MeasuredAI} for our method and baselines. \method variants achieve 16.3–39.4\% improvement in albedo estimation intensity, denoting highly improved robustness to over-exposure and generalization capability.}
\vspace{-1.5em}
\label{tab:maw}
\begin{center}
\scalebox{0.8}{
\begin{tabularx}{1.21\linewidth}{lcc}
\toprule
\textbf{Method} & \textbf{Intensity} \small{($\times$100)}\textbf{$~\downarrow$} & \textbf{Chromaticity~$\downarrow$} \\
\midrule
\citet{InteriorVerse}    & 1.44            & 4.94             \\
\citet{OrdinalShading}  & 0.57             & 6.56             \\
\citet{Kocsis2024IntrinsicID}    & 1.13       & 5.35    \\
\rgbx \cite{Zeng2024RGBXID}     & 0.82      & 3.96             \\
\citet{Careaga2024ColorfulDI}  & 0.54 & \cI 3.37    \\
OmniGen2 \cite{Wu2025OmniGen2ET} & \cI 0.41 & 4.40 \\
\midrule
Marigold (base variant)~\cite{Ke2025MarigoldAA} & \cIII 0.49 & 4.10 \\
\ours \marigoldx   & \cI \textbf{0.41}     &  \ours 4.01      \\
\multicolumn{1}{l}{\impr\textit{Improvement}} &\impr\textit{+16.3\%} & \impr\textit{+2.2\%} \\
\midrule
PRISM \cite{Dirik2025PRISMAU}   & 0.71     & \cIII 3.92        \\
\ours \prismx       & \cII \textbf{0.43}    & \cII \textbf{3.78}            \\
\multicolumn{1}{l}{\impr\textit{Improvement}} &\impr\textit{+39.4\%} & \impr\textit{+3.6\%} \\
\bottomrule
\end{tabularx}
}
\end{center}
\vspace{-1.5em}
\end{table}

As shown in \Cref{tab:iiw,tab:maw}, both \method variants (denoted \marigoldx and \prismx) significantly outperform their respective base models, with \prismx yielding SOTA zero-shot results on the IIW dataset that are comparable to CRefNet~\cite{luo2023crefnet}, a non-competing method trained on the IIW dataset. Similarly, \method variants achieve 16–39\% improvement over their base model on MAW intensity and achieve SOTA and comparable results, highlighting our method's effectiveness in improving decomposition robustness.

\textbf{Depth Estimation.} We separately evaluate the depth estimation quality of our method against a number of state-of-the-art monocular relative and metric depth estimation methods including Marigold Depth v1.0~\cite{ke2023repurposing} and Depth Anything V2 ~\cite{depth_anything_v1, depth_anything_v2}. We perform the comparisons on the NYUv2 \cite{Silberman2012IndoorSA} and ETH3D \cite{Schps2017AMS} datasets - two datasets neither our base model or \method variant have seen during training. To evaluate the depth estimation quality, we report Absolute Relative Error (AbsRel) and Threshold Accuracy ($\delta_1$) metrics. As \prism is trained to estimate disparity, we first convert PRISM and our \prismx estimated disparity maps to relative depth maps. As shown in Table \ref{tab:depth}, our \prismx model achieves SOTA or comparable results, as well as large improvements over its base PRISM model despite having been trained on real RGB images with no ground truth depth maps. We note that the improvement from the base model is more prominent on the ETH3D dataset, which is a predominantly outdoor dataset, whereas NYUv2 consists of indoor images.

\begin{table}[h]
\caption{Zero-shot relative depth estimation on NYUv2 and ETH3D datasets.  
Methods that require more than 2M samples for fine-tuning are denoted with gray. Our method achieves depth estimation results comparable to current SOTA methods despite having been trained on a fraction of the data, with our \prismx variant achieving 45.8\% AbsRel improvement over its base model on the mixed indoor / outdoor ETH3D dataset.}
\vspace{-1em}
\label{tab:depth}
\begin{center}
\scalebox{0.78}{%
\begin{tabularx}{1.29\linewidth}{lcccc}
\toprule
\textbf{Method} & \multicolumn{2}{c}{\textbf{NYUv2}} & \multicolumn{2}{c}{\textbf{ETH3D}} \\
\cmidrule(lr){2-3} \cmidrule(lr){4-5}
& \textbf{AbsRel $\downarrow$} & \textbf{$\delta_1$ $\uparrow$} & \textbf{AbsRel $\downarrow$} & \textbf{$\delta_1$ $\uparrow$} \\
\midrule
\textcolor{gray}{Omnidata}~\cite{eftekhar2021omnidata} & \cg 0.074 & \cg 0.945 & \cg 0.166 & \cg 0.778 \\
\textcolor{gray}{Depth Anything V2}~\cite{depth_anything_v2} & \cg 0.045 & \cg 0.979 & \cg 0.131 & \cg 0.865 \\
\midrule
MiDaS~\cite{Ranftl2019TowardsRM} & 0.111 & 0.885 & 0.184 & 0.752 \\
DPT~\cite{Ranftl2021} & 0.098 & 0.903 & \cIII 0.078 & \cIII 0.946 \\
HDN~\cite{zhang2022hierarchical} & 0.069 & \cIII 0.948 & 0.121 & 0.833 \\
OmniGen2 \cite{Wu2025OmniGen2ET}  & 0.111 & 0.886 & 0.123 & 0.897 \\
Marigold Depth \texttt{\small v1.0}~\cite{ke2023repurposing} & \cII 0.055 & \cI 0.964 & \cI 0.065 & \cI 0.960 \\
\midrule
PRISM~\cite{Dirik2025PRISMAU} & \cIII 0.061 & 0.922 & 0.142 & 0.836 \\
\ours \prismx & \cI \textbf{0.053} & \cII 0.958 & \cII 0.077 & \cII 0.950 \\
\multicolumn{1}{l}{\impr\textit{Improvement}} &\impr\textit{+13.1\%} & \impr\textit{+3.9\%} &\impr\textit{+45.8\%} &\impr \textit{+13.6\%} \\
\bottomrule
\end{tabularx}
}
\end{center}
\vspace{-1.5em}
\end{table}

\textbf{Surface Normals Estimation.}
We evaluate surface normal estimation performance of our \prismx model on the NYUv2 and DIODE \cite{diode_dataset} datasets and present quantitative comparisons with state-of-the-art methods in Table~\ref{tab:normals_results}. Our \prismx achieves the best mean angular error of 15.7° on NYUv2 and 14.5° on DIODE, demonstrating consistent improvements over the base PRISM model (16.1° and 14.6° respectively). While the improvements are modest compared to gains in other modalities, we argue that they demonstrate that our MLLM-guided approach can effectively refine geometric predictions even when the base model already performs competitively. Notably, our method outperforms recent specialized normal estimators including DSINE~\cite{Bae2024RethinkingIB}, GeoWizard~\cite{Fu2024GeoWizardUT}, and StableNormal~\cite{Ye2024StableNormalRD}, despite being trained without ground-truth normal supervision. We present qualitative depth and surface normals estimation comparisons with Marigold Depth v1.0, Marigold Normals v1.1 and the base PRISM model in Figures~\ref{fig:depth_comparison} and \ref{fig:normals_comparison}, where \prismx produces more accurate surface orientations, particularly in regions with complex geometry.

\begin{table}[h]
    \centering
    \caption{
        Zero-shot quantitative comparison of surface normals estimators on NYUv2 and DIODE datasets. Mean metric is in absolute angles, 11.25° metric in percentage. Our \prismx variant achieves consistently reports lower mean angular error compared to is base model PRISM, indicating increased robustness to edge cases.
    }
    \vspace{-0.5em}
    \scalebox{0.78}{
    \begin{tabularx}{1.3\linewidth}{lcccc}
        
        \toprule
        \textbf{Method} & \multicolumn{2}{c}{\textbf{NYUv2}} & \multicolumn{2}{c}{\textbf{DIODE}} \\
        
        &
        Mean ↓ & 
        11.25° ↑ &
        Mean ↓ & 
        11.25° ↑
        \\
        \midrule
        DSINE~\cite{Bae2024RethinkingIB} & \cIII 16.4 & \cII 59.6 & 19.9 &  41.8 \\
        GeoWizard~\cite{Fu2024GeoWizardUT} & 19.0 & 50.0 & 24.7 & 30.1 \\
        StableNormal~\cite{Ye2024StableNormalRD} & 17.8 & 54.2 & 19.3 & \cI 53.8 \\
        Lotus-G~\cite{He2024LotusDV} & 16.9 & \cIII 59.1 & 21.2 & 39.7 \\
        OmniGen2 \cite{Wu2025OmniGen2ET}  & 17.7 & 54.6 & 20.5 & 40.5 \\
        Marigold-Normals \texttt{\small v1.1}~\cite{Ke2025MarigoldAA} & \cII 16.1 & \cI 60.5 & \cIII 18.8 & \cII 45.5 \\
        
        \midrule
        PRISM ~\cite{Dirik2025PRISMAU} & \cII 16.1 & 56.8 & \cII 14.6 & \cIII 42.6 \\
        \ours \prismx & \cI \textbf{15.7} & \ours 57.0 & \cI \textbf{14.5} & 41.3 \\
        \multicolumn{1}{l}{\impr\textit{Improvement}} &\impr\textit{+2.5\%} &\impr \textit{+0.4\%} &\impr\textit{+0.7\%} &\impr \textit{–3.1\%} \\
        \bottomrule
        
    \end{tabularx}
     }
    \label{tab:normals_results}
\end{table}

\textbf{Effect on Cross Modal Alignment.}
A common issue when estimating each intrinsic channel separately is the alignment across different modalities, e.g., the white balance ambiguity between albedo and irradiance, alignment between depth and surface normals. We evaluate the effect of \method on alignment across modalities. In particular, we evaluate the alignment between depth and surface normal predictions by computing normals from estimated depth gradients and comparing to the estimated normals. We evaluate base model \prism and our \method variant, \prismx, on the ETH3D~\cite{Schps2017AMS} test set and 2000 samples from the COCO \cite{Lin2014MicrosoftCC} test set, and report the reconstruction metrics in \Cref{tab:depth-normals-alignment}. We exclude our \marigoldx variant from this experiment as both this variant and its base model (Marigold IID Lighting v.1.1) are trained to estimate albedo and irradiance only. As shown in \Cref{tab:depth-normals-alignment}, our \prismx achieves significant improvement over its base \prism on both datasets, highlighting the effectiveness of our method in improving geometric consistency.

\begin{table}
\caption{Given depth predictions on the ETH3D and COCO test sets, we compute the depth gradients and compare to the estimated normals to evaluate the alignment between the two channels.}
\vspace{-1.5em}
\label{tab:depth-normals-alignment}
\begin{center}
\scalebox{0.65}{%
\begin{tabularx}{1.54\linewidth}{llcccc}
\toprule
\textbf{Dataset} & \textbf{Method} & \textbf{RMSE $\downarrow$} & \textbf{PSNR $\uparrow$} & \textbf{SSIM $\uparrow$} & \textbf{LPIPS $\downarrow$} \\
\midrule
ETH3D & PRISM~\cite{Dirik2025PRISMAU} & 0.146 & 17.3 & 0.582 &  0.281 \\
 & \ours \prismx &\ours  0.099 \footnotesize (+32\%) & \ours 18.0 \footnotesize (+4.0\%) & \ours 0.640 \footnotesize (+10.0\%) & \ours 0.239 \footnotesize (+14.9\%) \\ 
\midrule
COCO & PRISM~\cite{Dirik2025PRISMAU} & 0.202 & 16.5 & 0.495 &  0.330 \\
 & \ours \prismx &\ours  0.137 \footnotesize (+32.2\%) & \ours 17.5 \footnotesize (+6.1\%) & \ours 0.579 \footnotesize (+17.0\%) & \ours 0.280 \footnotesize (+15.0\%) \\ 
\bottomrule
\end{tabularx}
}
\end{center}
\vspace{-2.5em}
\end{table}

\textbf{Qualitative Results.} We showcase \method qualitatively on a variety of challenging indoor and outdoor images in \Cref{fig:ood}. To evaluate the robustness under varying illumination, we test on the MIT Multi-Illumination Dataset (MIIW)~\cite{Murmann2019ADO}, which contains multiple images of each scene captured under different lighting conditions. Figure~\ref{fig:consistent-decomposition} shows decomposition results for the same scenes under two different illuminations using both base models (Marigold, PRISM) and their \method variants (\marigoldx, \prismx). The \method variants demonstrate significantly improved consistency across challenging lighting changes while base models produce noticeably different albedo estimates for the same scene under different illumination. We note that this improved disentanglement is particularly evident in regions with strong directional lighting or shadows, where \prismx and \marigoldx better preserve true material properties in the albedo channel, hence improving the albedo-irradiance alignment.

\subsection{Judge Accuracy and Validation}

\begin{figure}[t]
    \centering
    \includegraphics[width=1\linewidth]{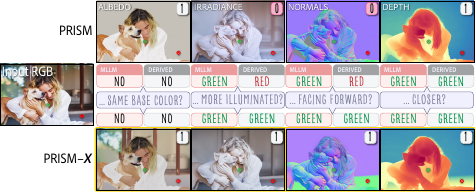}
    \vspace{-1em}
    \caption{Alignment of MLLM judgment and estimated intrinsics pre- and post-\method training of \prism.} 
    \label{fig:judge-pre-post-reasonx}
    \vspace{-1.0em}
\end{figure}

To assess the reliability of the intrinsic judge before applying it for policy fine-tuning, 
we evaluate its performance on the held-out test sets of InteriorVerse and HyperSim. 
Judge evaluation is inherently challenging, particularly for albedo and irradiance, where spatially extended markers and within-material color variation introduce visual ambiguity. 
Given an RGB image and its ground-truth intrinsic maps, we sample 20 point pairs and generate point-pair–annotated RGB images. 
For each $\{$RGB image, intrinsic image, point pair$\}$ triplet, the judge answers modality-specific questions, and we analytically derive the corresponding ground-truth judgment from the intrinsic image to compute the judge's accuracy and F1 score (\Cref{tab:judge-performance}). 
Because our visual markers cover small regions (not single pixels), the model must reason over textured neighborhoods rather than isolated points, making the notion of "same material color" or "same illumination" inherently ambiguous. 
Nevertheless, as shown in qualitative examples (Fig.~\ref{fig:judge-pre-post-reasonx}), the judge generally provides semantically correct feedback that improves global consistency in the fine-tuned model. 
Further implementation details and ablations on relative vs.\ absolute judgments are provided in the supplementary.

\begin{table}[h]
\caption{Our judge VLM performance on the held-out test set.}
\vspace{-1.5em}
\label{tab:judge-performance}
\begin{center}
\scalebox{0.8}{
\begin{tabularx}{1.1\linewidth}{lcccc}
\toprule
\textbf{Metric} & \textbf{Depth} & \textbf{Normal} & \textbf{Albedo} & \textbf{Irradiance} \\
\midrule
Accuracy        & 0.962 & 0.935 & 0.894 & 0.876 \\
Macro F1 score  & 0.962 & 0.933 & 0.889 & 0.878 \\
\bottomrule
\end{tabularx}
}
\end{center}
\vspace{-2.5em}
\end{table}

\section{Conclusion}
We introduced \method, a framework that uses relative intrinsic comparisons from an MLLM judge as rewards for GRPO-based fine-tuning of intrinsic decomposition models on unlabeled, in-the-wild images. By replacing absolute supervision with structured comparative feedback, \method enables effective fine-tuning of intrinsic decomposition models without requiring ground-truth intrinsic maps. Unlike prior GRPO uses with internal critics, our method leverages an external, MLLM-based reward model capable of cross-modal comparative reasoning. This enables supervision across intrinsic modalities without paired data. Our experiments show that \method consistently improves multiple base architectures and modalities, yielding substantial gains on real-world benchmarks—for example, a 25\% WHDR reduction on IIW and up to a 46\% depth accuracy improvement on ETH3D - despite the absence of annotated data during fine-tuning.

Limitations include the reliance on point-pair sampling and modality-wise rewards.  Hence, future work could explore joint multi-modal or reconstruction-based signals and extensions to broader inverse rendering tasks. Overall, \method demonstrates that MLLM-guided comparative supervision offers an effective pathway for bridging physical intrinsic reasoning with high-level perceptual understanding, providing a general paradigm for learning intrinsic decomposition models directly from real images.

\appendix
\appendix

\newpage
\section*{Appendix}
\section{Implementation}
For all experiments, we evaluate \method on two representative intrinsic decomposition architectures: (i) PRISM~\cite{Dirik2025PRISMAU}, a rectified-flow diffusion transformer for RGBX prediction, and (ii) Marigold IID Lighting v1.1~\cite{Ke2025MarigoldAA}, a diffusion-based model for joint albedo-irradiance estimation. We fine-tune both models using our Intrinsic-GRPO framework, with the MLLM judge providing relative intrinsic comparison rewards. Unless otherwise stated, all fine-tuning follows the setup described in the main paper: a training set of 10k COCO~\cite{Lin2014MicrosoftCC} images, $T{=}15$ denoising steps ($T{=}50$ steps during inference), AdamW optimizer with a learning rate of $10^{-5}$, group size $G{=}8$, and SDE noise scale $a{=}0.7$. PRISM is conditioned using an empty text prompt in addition to the RGB image condition at all steps during training and post-training / \prismx inference.

For evaluation, we compare our \method variants against their base models and a diverse set of strong supervised and synthetic-data methods. Additionally, we evaluate against OmniGen2~\cite{Wu2025OmniGen2ET}, where we follow the fine-tuning strategy described in the main paper to adapt its unified generation framework for multi-task intrinsic prediction. We evaluate all baselines using their publicly released models, official code bases, and default / highest performing settings.

\section{MLLM Judge}
\subsection{Training Details}
We use InternVL2.5-4B~\cite{Chen2024ExpandingPB} as our base model for our MLLM judge and fine-tune it on (point-pair annotated RGB image, ground truth judgment) pairs. For training and evaluation, we use the same training / test split of the synthetic HyperSim \cite{roberts2021hypersim} and InteriorVerse \cite{InteriorVerse} datasets as our base model \prism, use an image resolution of $512\times512$px and fine-tune the base MLLM for 5 epochs. For point-pair sampling, we set a minimum and maximum distance threshold of 20 and 350 between the points, and discard point pairs with less than 2\% difference between them. 

\subsection{Absolute vs Relative Intrinsic Judgments}
To assess whether relative comparisons are advantageous over absolute predictions, and whether such relative comparisons can be derived from absolute intrinsic predictions, we conduct an ablation study in which the base MLLM is fine-tuned to predict discretized absolute intrinsic values. For each modality, we partition the continuous ground-truth range into five ordered bins (e.g. for depth: very near, near, mid, far, very far; for albedo: very dark to very bright). We fine-tune separate MLLMs for each modality to perform 5-way classification on single annotated points, using the same synthetic training and test split as our relative judge. At test time, we evaluate both absolute accuracy and the quality of \emph{derived} relative judgments by comparing the predicted bins for a pair of points and inferring their ordinal relationship. We exclude point pairs where both points belong to the same bin for simplicity.

\begin{table}[h]
\caption{Performance of MLLM fine-tuned to perform absolute predictions (5-bin). We report accuracy and F1 score metrics for both the absolute predictions and relative judgments derived from them.}
\vspace{-1.5em}
\label{tab:ablation-performance}
\begin{center}
\scalebox{0.9}{
\begin{tabularx}{1.1\linewidth}{lcccc}
\toprule
\textbf{Modality} & \multicolumn{2}{c}{\textbf{Absolute}} & \multicolumn{2}{c}{\textbf{Relative (Derived)}} \\
& \textbf{Accuracy} & \textbf{F1 Score} & \textbf{Accuracy} & \textbf{Macro F1} \\
\midrule
Depth        & 0.634 & 0.401 & 0.475 & 0.451 \\
Normal       & 0.800 & 0.265 & 0.702 & 0.690 \\
Albedo       & 0.449 & 0.426 & 0.483 & 0.488 \\
Irradiance   & 0.541 & 0.437 & 0.418 & 0.387 \\
\bottomrule
\end{tabularx}
}
\end{center}
\vspace{-2em}
\end{table}

As shown in Table \ref{tab:ablation-performance}, absolute predictions prove substantially more challenging for the MLLM, with accuracies ranging from 44.9\% to 80.0\% depending on the modality. Moreover, relative judgments obtained by differencing these absolute predictions perform significantly worse than those from our directly trained relative judge. For example, derived depth comparisons achieve only 47.5\% accuracy compared to 96.2\% with our relative model. These results validate our relative judgment approach as both more effective and better suited to the comparative nature of the task.
\vspace{-.5em}

\begin{figure*}[t]
    \centering
    \includegraphics[width=1\linewidth]{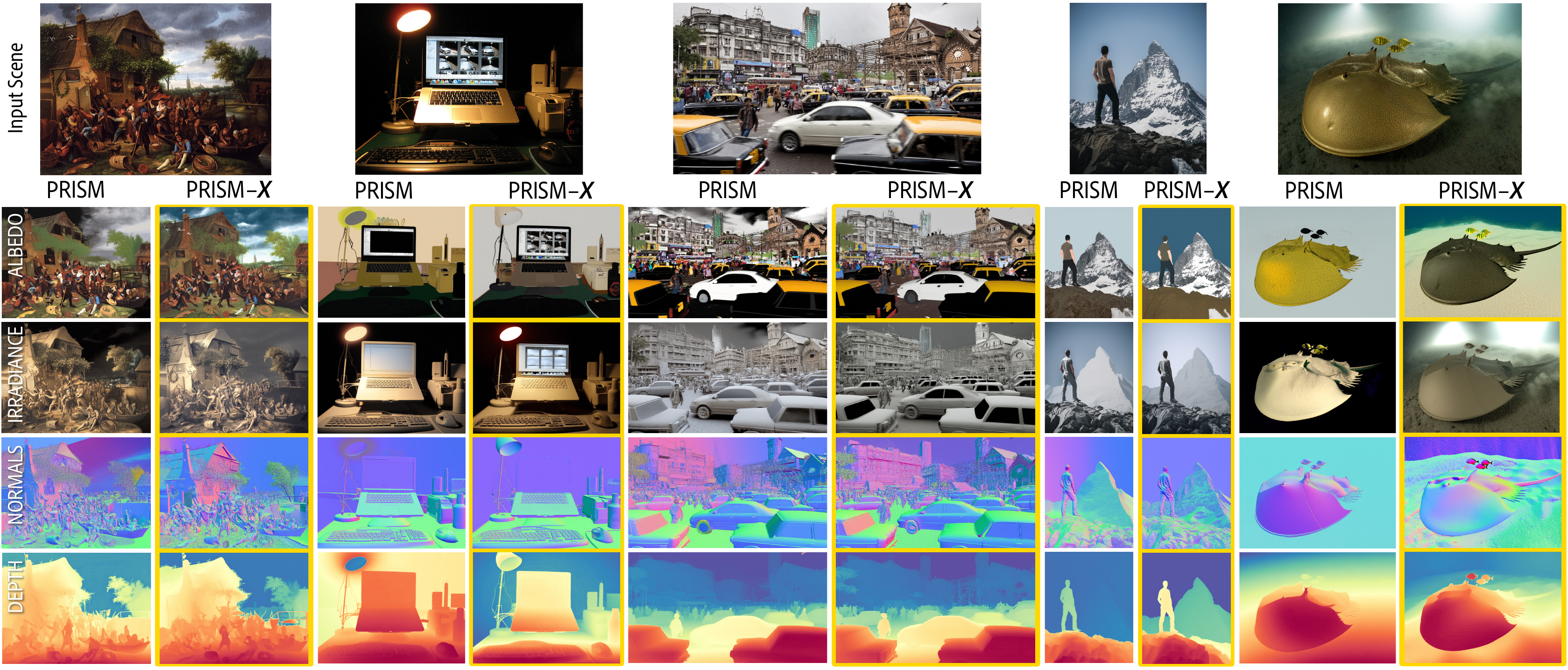}
    \vspace{-2em}
    \caption{Additional intrinsic decomposition samples on challenging out-of-distribution images. Our PRISM-\textbf{X} significantly improves its base model PRISM across all intrinsic channels with respect to decomposition quality and in-the-wild generalization performance.} 
    \label{fig:ood-supp-prism}
    \vspace{-.5em}
\end{figure*}

\subsection{Effect of \method on Alignment Rewards}
\begin{figure*}[h]
    \centering
    \includegraphics[width=1\linewidth]{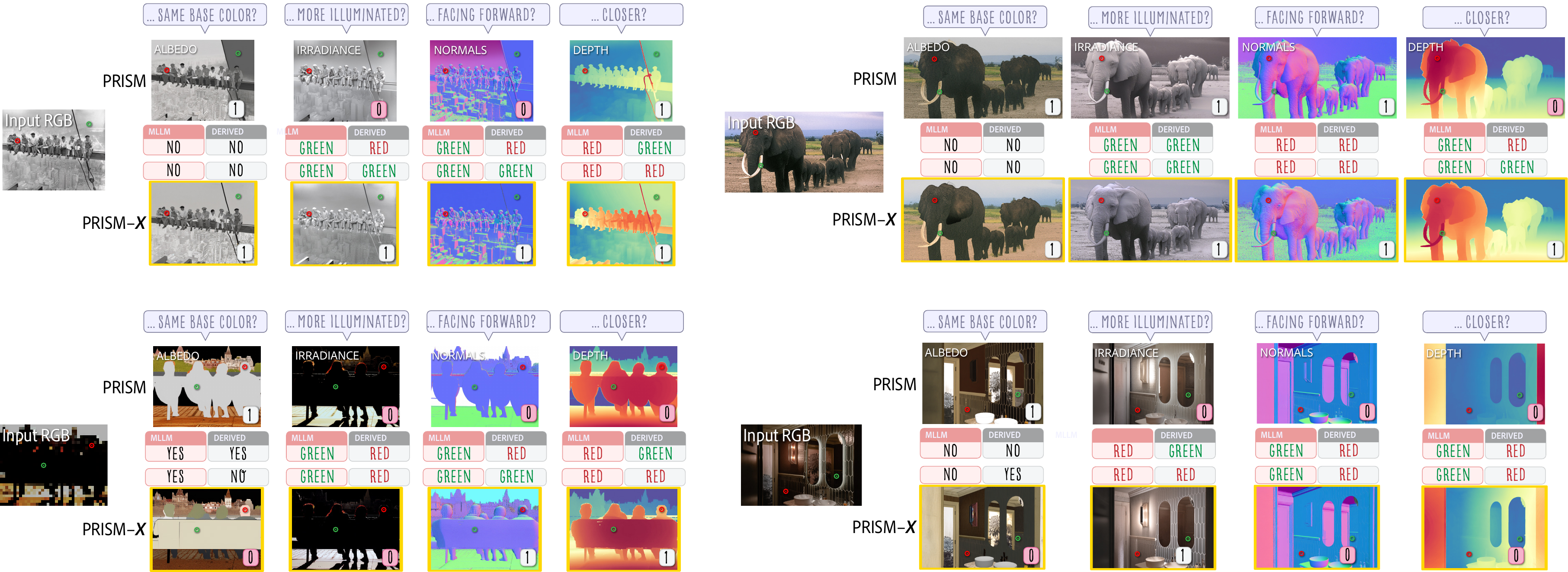}
    \vspace{-2em}
    \caption{Alignment of MLLM judgments with analytically derived judgments from predicted intrinsics pre and post \method training of base model \prism. Top rows shows samples with accurate MLLM judgments where \prismx generated intrinsics result in improved alignment. Bottom row show failure cases where both the MLLM and analytically derived judgments might be unreliable due to dim lighting, intricate patterns and mirrored / reflective surfaces. } 
    \label{fig:pre-post-reasonx-supp}
    \vspace{-1em}
\end{figure*}

A central goal of \method is to improve the agreement between (i) the MLLM judge’s relative intrinsic judgments and (ii) the analytic relations implied by the model’s predicted intrinsics. To illustrate this effect, we visualize the alignment before and after GRPO finetuning (Fig.~\ref{fig:pre-post-reasonx-supp}). For each example, we show the judge MLLM’s responses to point-pair annotated RGB image queries per modality-specific question alongside the corresponding analytic comparisons computed from the predicted intrinsics. In the top rows, \method produces noticeably stronger alignment: the improved intrinsic predictions yield analytic relations that match the judge’s comparative assessments more consistently. The bottom rows highlight challenging cases, such as highly textured materials (e.g. marble, wood), intricate patterns, mirrors and reflective surfaces, and scenes with dim or spatially complex lighting) where both the intrinsic predictions and the judge’s assessments may be unreliable. These examples illustrate the inherent ambiguity of such scenes and mark natural failure modes shared by both analytic and perceptual signals. Overall, the pre/post \method comparison supports that \method training strengthens cross-modal consistency and enhances the agreement between analytic intrinsic relations and MLLM-based comparative reasoning.
\vspace{-1em}

\begin{figure*}[t]
    \centering
    \includegraphics[width=1\linewidth]{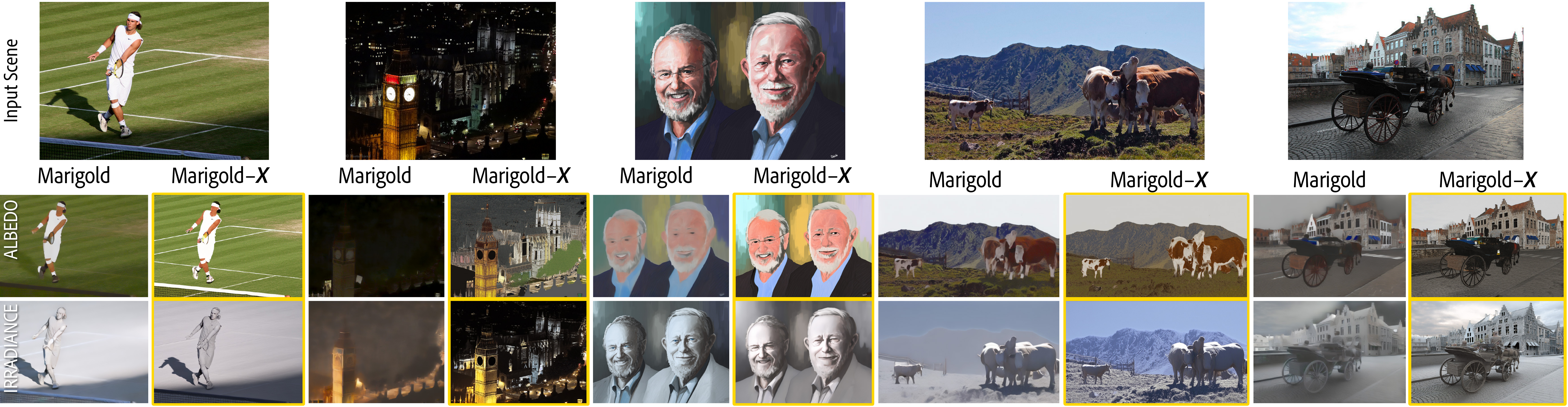}
    \vspace{-2em}
    \caption{Intrinsic decomposition on challenging out-of-distribution images with the base Marigold (IID Lightning v1.1) model and our Marigold-\textbf{X}. Our \method variant significantly improves its base model Marigold, especially in cases where the true albedo / base color is difficult to interpret.} 
    \label{fig:ood-supp-marigold}
    \vspace{-.5em}
\end{figure*}

\section{Additional Results}
We provide further quantitative and qualitative evaluations to complement the results in the main paper. First, we assess the generalization of \method variants on the unseen synthetic ARAP dataset~\cite{BKPB17}, following the evaluation protocol of CID~\cite{Careaga2024ColorfulDI}. As shown in \Cref{tab:arap}, our \prismx and \marigoldx variants demonstrate improvements over the base PRISM and Marigold models as well as comparisons to strong baselines. Across all metrics, \method yields consistent gains and achieves state-of-the-art performance.

We then evaluate our \method variants on the synthetic HyperSim and InteriorVerse test sets against several strong baselines, and present our results in Tables~\ref{tab:hypersim} and \ref{tab:interiorverse}. On both datasets, \method consistently improves the quality of albedo, normals, and irradiance predictions. Finally, we include extensive qualitative comparisons for both \prismx and \marigoldx on challenging in-the-wild and out-of-distribution images (Figs.~\ref{fig:ood-supp-prism} and~\ref{fig:ood-supp-marigold}), illustrating the improved decomposition quality and cross-modal consistency achieved by our framework.

\begin{table}[h]
\caption{Average MLLM judgment - predicted intrinsics alignment rewards pre- and post-optimization of our base models on our COCO test set.}
\vspace{-1.5em}
\label{tab:pre-post-rewards}
\begin{center}
\scalebox{0.78}{%
\begin{tabularx}{1.31\linewidth}{lcccc}
\toprule
\textbf{Method} & \textbf{Albedo} & \textbf{Irradiance} & \textbf{Normals} & \textbf{Depth} \\
\toprule
Marigold (base variant)~\cite{ke2023repurposing} & 0.735 & 0.588 & - & - \\
\ours \marigoldx & \ours 0.801 & \ours  0.694 & - & - \\
\midrule
PRISM~\cite{Dirik2025PRISMAU} & 0.701 & 0.597 & 0.618 &  0.591 \\
\ours \prismx &\ours  0.789 & \ours 0.708 & \ours 0.701 & \ours 0.771 \\ 
\bottomrule
\end{tabularx}
}
\end{center}
\vspace{-2em}
\end{table}

\begin{table}[h]
\vspace{-0.5em}
\caption{Albedo decomposition quality evaluation on the synthetic ARAP Dataset~\cite{BKPB17}. Our \method variants achieve 4.7–5.7\% RMSE improvement over their base models.
}
\vspace{-1.5em}
\label{tab:arap}
\begin{center}
\scalebox{0.8}{
\begin{tabularx}{1.03\linewidth}{lcccc} 
\toprule
\renewcommand{\arraystretch}{0.9}
\textbf{Method} & \textbf{LMSE}~$\downarrow$ & \textbf{RMSE}~$\downarrow$ & \textbf{SSIM}~$\uparrow$ \\
\midrule
\citet{InteriorVerse} & 0.029 & 0.184 & 0.729 \\
\citet{OrdinalShading} & 0.035 & 0.162 & 0.751 \\
\midrule
\citet{Kocsis2024IntrinsicID} & 0.030 & 0.160 & 0.738 \\
\rgbx \cite{Zeng2024RGBXID}  & 0.025 & 0.177 & 0.738      \\
\citet{Careaga2024ColorfulDI} & \cII 0.021 & \cII 0.149 & \cIII 0.796 \\
OmniGen2 \cite{Wu2025OmniGen2ET} & 0.046 & 0.225 & 0.668 \\
\midrule
Marigold \cite{Ke2025MarigoldAA} & \cIII 0.022 & \cIII 0.158 & 0.732 \\
\ours \marigoldx & \cII 0.021 & \ours \cII 0.149 &  \ours 0.781\\
\multicolumn{1}{l}{\impr\textit{Improvement}} &\impr \textit{+4.5\%} &\impr \textit{+5.7\%} & \impr\textit{+6.7\%} \\
\midrule
PRISM \cite{Dirik2025PRISMAU} & \cIII 0.022 & \cII 0.149 & \cII 0.798 \\
\ours \prismx &\cI $\mathbf{0.020}$ & \cI $\mathbf{0.142}$  &  \cI $\mathbf{0.820}$ \\
\multicolumn{1}{l}{\impr\textit{Improvement}} &\impr\textit{+9.1\%} & \impr\textit{+4.7\%} & \impr\textit{+2.8\%} \\
\bottomrule
\end{tabularx}
}
\end{center}
\vspace{-1em}
\end{table}

\begin{table}[h]  
    \caption{
        Quantitative evaluation of our method's intrinsic image decomposition performance on the HyperSim test set against various baselines.
    }
    \vspace{-1em}
    \label{tab:hypersim}
    \scalebox{0.63}{
        \begin{tabularx}{1.61\linewidth}{ccccccc}
            \toprule
             \multirow{3}{*}{\textbf{Method}} & \multicolumn{2}{c}{\textbf{Albedo}} & \multicolumn{2}{c}{\textbf{Normal}} & \multicolumn{2}{c}{\textbf{Irradiance}}                                          \\
            \cmidrule(lr){2-3}
            \cmidrule(lr){4-5}
            \cmidrule(lr){6-7}
             \multicolumn{1}{c}{} & {\small\textbf{PSNR}~$\uparrow$} & {\small\textbf{LPIPS}~$\downarrow$} & {\small\textbf{PSNR}~$\uparrow$} & {\small\textbf{LPIPS}~$\downarrow$} & {\small\textbf{PSNR}~$\uparrow$} & {\small\textbf{LPIPS}~$\downarrow$} 
            \\
            \midrule
             \citet{InteriorVerse} & 11.7 & 0.54 & 16.5 & 0.45 & -- & --        \\
             \citet{OrdinalShading} & 13.5 & 0.34 & -- & -- & 14.5 & 0.22 \\
             \citet{Kocsis2024IntrinsicID} & 12.1 & 0.41 & -- & -- & -- & -- \\
             \rgbx \cite{Zeng2024RGBXID} & 17.4 & \cII 0.18 & \cIII 19.8 & \cII 0.18 & 14.1 & 0.22 \\
             \citet{Careaga2024ColorfulDI} & 17.1 & 0.21 & -- & -- & 17.2 & 0.21 \\

             OmniGen2 \cite{Wu2025OmniGen2ET} & 17.6 & \cIII 0.19 & 19.5 & 0.21 & 16.5 & \cII 0.19 \\
             Marigold-Normals v1.1 \cite{Ke2025MarigoldAA} & -- & -- & \cIII 19.8 & \cIII 0.20 & -- & -- \\
            \midrule
             Marigold \cite{Ke2025MarigoldAA} & 18.2 & 0.22 & -- & -- & 17.6 & 0.26 \\
            
             \ours \marigoldx & \cIII 19.0 & 0.20 & -- & -- & \cIII 18.2 &  0.21 \\
            \midrule
             PRISM \cite{Dirik2025PRISMAU} & \cII 19.3 & \cII 0.18 & \cII 19.9 & \cII 0.18 & \cII 18.5 & \cIII 0.20 \\
            
             \ours \prismx & \cI \textbf{19.9} & \cI \textbf{0.17} & \cI \textbf{20.3} & \cI \textbf{0.17} & \cI \textbf{18.9} & \cI \textbf{0.18} \\
            \bottomrule
        \end{tabularx}
        }
\vspace{-1em}
\end{table}

\begin{table}[h]  
    \caption{
        Quantitative evaluation of our method's intrinsic image decomposition performance on the InteriorVerse test set against various baselines.
    }
    \vspace{-1em}
    \label{tab:interiorverse}
    \centering
    \scalebox{0.7}{
        \begin{tabularx}{1.23\linewidth}{ccccc}
            \toprule
            \multirow{3}{*}{\textbf{Method}} & \multicolumn{2}{c}{\textbf{Albedo}} & \multicolumn{2}{c}{\textbf{Normal}}                            \\
            \cmidrule(lr){2-3}
            \cmidrule(lr){4-5}
            \multicolumn{1}{c}{} & {\small\textbf{PSNR}~$\uparrow$} & {\small\textbf{LPIPS}~$\downarrow$} & {\small\textbf{PSNR}~$\uparrow$} & {\small\textbf{LPIPS}~$\downarrow$}  
            \\
            \midrule
             \citet{InteriorVerse} & 13.6 & 0.24 & 17.1 & 0.26  \\
             \citet{OrdinalShading} & 17.4 & 0.20 & -- & --  \\
             \citet{Kocsis2024IntrinsicID} & 12.2 & 0.30 & 20.1 & 0.21  \\
             \rgbx \cite{Zeng2024RGBXID}  & 16.6 & 0.17 & \cIII 20.2 & 0.19 \\
             \citet{Careaga2024ColorfulDI} & 17.7 & 0.27 & -- & --  \\
             OmniGen2 \cite{Wu2025OmniGen2ET} & 13.9 & 0.21 & 19.0 & 0.18 \\
             Marigold-Normals v1.1\cite{Ke2025MarigoldAA} & -- & -- & \cIII 20.2 & \cIII 0.16 \\
            \midrule
             Marigold \cite{Ke2025MarigoldAA} & 19.5 & 0.19 & -- & --  \\
             \ours \marigoldx & \cIII 19.8 & \cIII 0.16 & -- & --  \\
            \midrule
             PRISM \cite{Dirik2025PRISMAU} & \cII 19.9 & \cII 0.14 & \cII 21.2 & \cII  0.15 \\
             \ours \prismx & \cI \textbf{20.7} & \cI \textbf{0.12} & \cI \textbf{21.5} & \cI \textbf{0.14}  \\
            \bottomrule
        \end{tabularx}
        }
\vspace{-.5em}
\end{table}

\subsection{Cross-Modal Alignment} 
We perform a cyclic RGB reconstruction experiment on a subset of the COCO test set \cite{Lin2014MicrosoftCC} consisting of 5000 images to evaluate the effect of \method on the alignment between the estimated modalities. Given a source RGB image, we reconstruct the diffuse appearance of the target image simply by multiplying the estimated albedo and irradiance maps. For our Marigold base model and \method variant, we additionally apply a shift using the Marigold predicted residual ($RGB = A*I + R$). We then compare the reconstructed image to the source image and report the reconstruction metrics in \Cref{tab:rgb-to-x-to-rgb}. As shown in Table~\ref{tab:rgb-to-x-to-rgb}, both \method variants exhibit improved cyclic reconstruction quality over their base models, indicating that our \method framework not only improves individual modalities but also improves their cross-modal alignment. We note that these improvements are strongly correlated with more stable and accurate albedo estimation.

\begin{table}
\caption{Cyclic RGB (RGB-to-X-to-RGB) reconstruction error on the real COCO test set. We perform intrinsic decomposition with our \method variants and reconstruct RGB images using the estimation albedo and irradiance. Our \method variants show improved reconstruction over their base models, indicating better cross-alignment of modalities.}
\vspace{-1.5em}
\label{tab:rgb-to-x-to-rgb}
\begin{center}
\scalebox{0.78}{%
\begin{tabularx}{1.29\linewidth}{lcccc}
\toprule
\textbf{Method} & \textbf{RMSE $\downarrow$} & \textbf{PSNR $\uparrow$} & \textbf{SSIM $\uparrow$} & \textbf{LPIPS $\downarrow$} \\
\toprule
Marigold (base variant)~\cite{ke2023repurposing} & 0.184 & 14.7 & 0.564 & 0.35 \\
\ours \marigoldx & \ours 0.178 & \ours  15.0 &\ours  0.581 &\ours  0.32 \\
\multicolumn{1}{l}{\impr\textit{Improvement}} &\impr\textit{+3.3\%} &\impr \textit{+2.0\%} &\impr\textit{+3.0\%} & \impr\textit{+8.6\%} \\
\midrule
PRISM~\cite{Dirik2025PRISMAU} & 0.192 & 14.3 & 0.584 &  0.30 \\
\ours \prismx &\ours  0.184 & \ours 14.7 & \ours 0.610 & \ours 0.28 \\ 
\multicolumn{1}{l}{\impr\textit{Improvement}} &\impr\textit{+4.2\%} & \impr\textit{+2.8\%} &\impr\textit{+6.7\%} & \impr\textit{–3.1\%} \\
\bottomrule
\end{tabularx}
}
\end{center}
\end{table}

\section{Ablation Studies}
\subsection{Analytical Depth-Normal Consistency Reward}
To assess the contribution of the MLLM judge, we perform an ablation study in which the MLLM-based comparative reward used in our GRPO loop is replaced with a purely analytic geometric reward based on the consistency of the predicted depth and normals. We note that this ablation leverages the fact that \prism predicts multiple intrinsic modalities jointly, making this type of
consistency reward generally unavailable to intrinsic decomposition models. Given \prism predicted depth $\hat{D}$ and normals $\hat{N}$, we estimate a depth map $\tilde{D}$ implied by the predicted normals using a least-squares Poisson integration. The reward penalizes disagreement between  
the two depth maps: $r_{\mathrm{DN}} = -\|\hat{D} - \tilde{D}\|_{1}.$

This consistency signal enforces geometric coherence but provides no explicit supervision on albedo or irradiance, and does not incorporate perceptual cues. We finetune \method using this analytic reward under the same GRPO setup as our main experiments and present the results in Table~\ref{tab:dn-ablation}. While the depth-normal reward improves geometric consistency, it yields smaller gains than the full \method framework and does not transfer improvements to other modalities (e.g., albedo). These findings highlight that purely analytic self-consistency is insufficient for robust improvement on real images, whereas MLLM-derived relative feedback offers a richer and more transferable signal.

\begin{table}[h]
\caption{
Depth and normals ablation using a depth-normal consistency (DN) reward. DN improves over the base \prism model but is consistently weaker than \method. 
}
\vspace{-0.75em}
\label{tab:dn-ablation}
\centering
\scalebox{0.7}{
\begin{tabular}{lcccccc}
\toprule
& \multicolumn{2}{c}{\textbf{NYU-v2}} & \multicolumn{2}{c}{\textbf{ETH3D}} & \multicolumn{2}{c}{\textbf{DIODE}} \\
\cmidrule(lr){2-3} \cmidrule(lr){4-5} \cmidrule(lr){6-7}
\textbf{Method} & AbsRel↓ & $\delta_1$↑ & AbsRel↓ & $\delta_1$↑ & Mean↓ & $<\!11.5^\circ$↑ \\
\midrule
\prism              & 0.061 & 0.922 & 0.142 & 0.836 & 14.6 & 42.6  \\
\prism + DN reward  & 0.060 & 0.930 & 0.089 & 0.905 & 14.8 & \textbf{43.0} \\
\ours \prismx             &\ours \textbf{0.053} &\ours \textbf{0.958} &
                       \ours \textbf{0.077} & \ours \textbf{0.950} & \ours \textbf{14.5} & \ours 41.3\\
\bottomrule
\end{tabular}}
\end{table}

\subsection{Effect of KL Regularization}
We use a KL regularization term during GRPO to prevent reward hacking and overly aggressive updates that drift from the pretrained model distribution~\cite{Liu2025FlowGRPOTF}. To isolate its contribution, we consider a variant of \method in which the KL term is removed from the GRPO objective. As such a setup leads to rapid model drift, we adopt an interleaved training strategy where we perform GRPO on real images without KL regularization, followed by standard flow-matching updates on the original synthetic training set of \prism. This alternating procedure stabilizes optimization while allowing us to assess the effect of removing KL regularization. We report results on synthetic intrinsic benchmarks and depth estimation datasets in Tables~\ref{tab:ablation_kl_synthetic} and \ref{tab:ablation_kl_depth} respectively. 

\begin{table}[h]
\caption{Ablation study: Impact of KL regularization on albedo, normals and irradiance decomposition across synthetic datasets.}
\label{tab:ablation_kl_synthetic}
\vspace{-.5em}
\scalebox{0.75}{%
\begin{tabularx}{1.3\linewidth}{lcccccc}
\toprule
\multirow{2}{*}{\textbf{Method}} & \multicolumn{2}{c}{\textbf{Albedo}} & \multicolumn{2}{c}{\textbf{Normal}} & \multicolumn{2}{c}{\textbf{Irradiance}} \\
\cmidrule(lr){2-3} \cmidrule(lr){4-5} \cmidrule(lr){6-7}
& PSNR↑ & LPIPS↓ & PSNR↑ & LPIPS↓ & PSNR↑ & LPIPS↓ \\
\midrule
\multicolumn{7}{c}{\textit{HyperSim}} \\
\ours \prismx & \ours 19.9 & \ours 0.17 & \ours 20.3 & \ours 0.17 & \ours 18.9 & \ours 0.18 \\
w/o KL reg. & 19.5 & 0.18 & 20.0 & 0.18 & 18.6 & 0.19 \\
\midrule
\multicolumn{7}{c}{\textit{InteriorVerse}} \\
\ours \prismx & \ours 20.7 & \ours 0.12 & \ours 21.5 & \ours 0.14 & - & - \\
w/o KL reg. & 20.1 & 0.14 & 21.2 & 0.15 & - & - \\
\bottomrule
\end{tabularx}
}
\end{table}

\begin{table}[h]
\caption{Ablation study: Impact of KL regularization on depth estimation.}
\label{tab:ablation_kl_depth}
\vspace{-1.5em}
\begin{center}
\scalebox{0.9}{%
\begin{tabularx}{\linewidth}{lcccc}
\toprule
\multirow{2}{*}{\textbf{Method}} & \multicolumn{2}{c}{\textbf{NYU-v2}} & \multicolumn{2}{c}{\textbf{ETH3D}} \\
\cmidrule(lr){2-3} \cmidrule(lr){4-5}
& AbsRel↓ & $\delta_1$↑ & AbsRel↓ & $\delta_1$↑ \\
\midrule
\ours \prismx & \ours 0.053 & \ours 0.958 & \ours 0.077 & \ours 0.950 \\
w/o KL reg. & 0.059 & 0.930 & 0.135 & 0.864 \\
\bottomrule
\end{tabularx}
}
\end{center}
\vspace{-.5em}
\end{table}

As shown in Tables~\ref{tab:ablation_kl_synthetic} and \ref{tab:ablation_kl_depth}, despite the interleaved stabilization strategy, removing the KL term yields consistent degradation across albedo, normals, irradiance, and depth. The drop is most pronounced on ETH3D, a mixed indoor-outdoor dataset, which indicates reduced generalization capabilities compared to our \prismx model. These results highlight that KL regularization is required for stable optimization under comparative rewards, ensuring that improvements arise from meaningful alignment with the judge rather than from exploiting the reward structure or drifting away from the pretrained model manifold.

{
\small
\bibliographystyle{ieeenat_fullname}
\bibliography{main}
}

{
\small
\bibliographystyle{ieeenat_fullname}
\bibliography{main}
}

\end{document}